# A Multi-Policy Framework for Deep Learning-Based Fake News Detection

João Vitorino[12][0000-0002-4968-3653], Tiago Dias[12][0000-0002-1693-7872], Tiago Fonseca[1][0000-0002-5592-3107], Nuno Oliveira[12][0000-0002-5030-7751] and Isabel Praça[12][0000-0002-2519-9859]

[1] School of Engineering, Polytechnic of Porto (ISEP/IPP), 4249-015 Porto, Portugal
[2] Research Group on Intelligent Engineering and Computing for Advanced Innovation and Development (GECAD), 4249-015 Porto, Portugal
{jpmvo,tiada,calof,nunal,icp}@isep.ipp.pt

**Abstract.** Connectivity plays an ever-increasing role in modern society, with people all around the world having easy access to rapidly disseminated information. However, a more interconnected society enables the spread of intentionally false information. To mitigate the negative impacts of fake news, it is essential to improve detection methodologies. This work introduces Multi-Policy Statement Checker (MPSC), a framework that automates fake news detection by using deep learning techniques to analyze a statement itself and its related news articles, predicting whether it is seemingly credible or suspicious. The proposed framework was evaluated using four merged datasets containing real and fake news. Long-Short Term Memory (LSTM), Gated Recurrent Unit (GRU) and Bidirectional Encoder Representations from Transformers (BERT) models were trained to utilize both lexical and syntactic features, and their performance was evaluated. The obtained results demonstrate that a multi-policy analysis reliably identifies suspicious statements, which can be advantageous for fake news detection.

## 1 Introduction

Fake news are statements that are intentionally and verifiably false. They can be used for disinformation attacks, attempting to manipulate society's perception of real facts and events [1]. Exposure to fake news can have significantly negative impacts in various fields, being especially concerning in healthcare [2] and politics [3]. This is considered an emerging cybersecurity threat because it is one of the main weapons of information warfare and can even lead to real-world risks to public safety.

Although the cybersecurity threat that fake news pose to modern society is not a novelty, the rise of social networking platforms and messaging applications has led to alarming dissemination rates. Despite being developed to improve connectivity and simplify access to information, the wide reach of these platforms can be exploited to perform disinformation attacks [4]. Therefore, it is of the utmost importance nowadays to identify false information and mitigate its impact.

The traditional methodology for verifying information is assigning professional journalists to investigate claims and fact-check them against concrete evidence. Nonetheless, this process is both expensive and time-consuming. To efficiently process large amounts of widespread information, automated fake news detection is required. Artificial intelligence is at the forefront of this task because it has the potential to significantly improve Natural Language Processing (NLP) tasks.

This work addresses the detection of suspicious information, introducing the Multi-Policy Statement Checker (MPSC) to analyze the legitimacy of a statement. Deep learning techniques are utilized to predict whether a statement is seemingly credible or suspicious, according to both lexical and syntactic features that are obtained from the statement itself and from related news articles. The performance of the proposed framework was evaluated using four merged fake news detection datasets and by training Long-Short Term Memory (LSTM), Gated Recurrent Unit (GRU) and Bidirectional Encoder Representations from Transformers (BERT) models. Additionally, a visual representation of word frequency in the credible and suspicious statements of the merged dataset was also provided.

The present paper is organized into multiple sections. Section 2 provides a survey of previous work on fake news detection. Section 3 describes the proposed framework and the concepts it relies on. Section 4 presents the case study and an analysis of the obtained results. Finally, Section 5 addresses the main conclusions and future work.

## 2 Related Work

In recent years, fake news detection has been gaining relevance as a research topic. As novel NLP techniques are developed and computational resources are increased, more reliable approaches are being developed to analyze information. However, despite deceitful statements usually incurring grammatical errors or using strong expressions, it is only possible to distinguish between real and fake information by analyzing their surrounding context [5].

Shu et al. [6] address the detection of fake social media posts from a data mining perspective, characterizing psychology and social theories, as well as examining existing data mining algorithms, evaluation metrics and representative datasets. Additionally, Mridha et al. [2] review state-of-the-art techniques for data preprocessing, word vectorizing and feature extraction, while also detailing current deep learning techniques for fake news detection, such as Convolutional Neural Networks (CNN), Recurrent Neural Network (RNN), Generative Adversarial Network (GAN) and BERT. To benefit from the advantages of the latter technique, Kaliyar et al. [7] propose a BERT-based deep learning sentence encoder approach capable of extracting the context representation of a sentence, in combination with single-layer deep CNN capturing both semantic and long-distance relationships between sentences.

Reis et al. in [8] introduced a supervised learning approach to automatic fake news detection with a new set of features. The authors extracted lexical and psycholinguis-tic features, as well as media engagement, source trustworthiness and location, and evaluated their discriminative classification capability. Even though nearly all fake news could be detected, 40% of true news were misclassified as fake news.

Several authors have also considered multi-modal classification, mainly focused on simultaneously analyzing text and images [9]–[11]. Shivangi et al. [12] proposed a multi-modal approach for the classification of news articles, combining text and images. However, it would require an accompanying image for each statement, hindering the analysis of individual statements. On the other hand, content verification approaches have achieved reliable performances [13]–[16]. For instance, Ibrishimova and Li [17] introduced a framework that performed a syntactic analysis and attempted to check the veracity of a statement. The computed knowledge verification features could be used to correctly identify suspicious news, although they did not benefit from the full lexical analysis of well-established techniques.

To address the drawbacks of the current literature, this work introduces multiple policies to be followed for fake news detection. To the best of our knowledge, no previous work has introduced a similar approach.

## 3 Proposed Framework

MPSC was developed with the objective of automating the detection of false information, so large amounts of information can be filtered and professional fact-checking can be performed on a small number of statements. Therefore, it was pertinent to analyze both a statement and its context, predicting its level of suspiciousness. The following subsections detail the considered policies and the architecture of the analysis performed within the framework.

### 3.1 Considered Policies

To account for a statement's context, it is necessary to also analyze the news articles related to its main subjects. Two policies are systematized: processing the information of a statement itself and processing the information obtained from a news search (see Fig. 1). The statement and the contents of the obtained articles are then combined in a deep learning analysis step to perform a multi-policy classification.

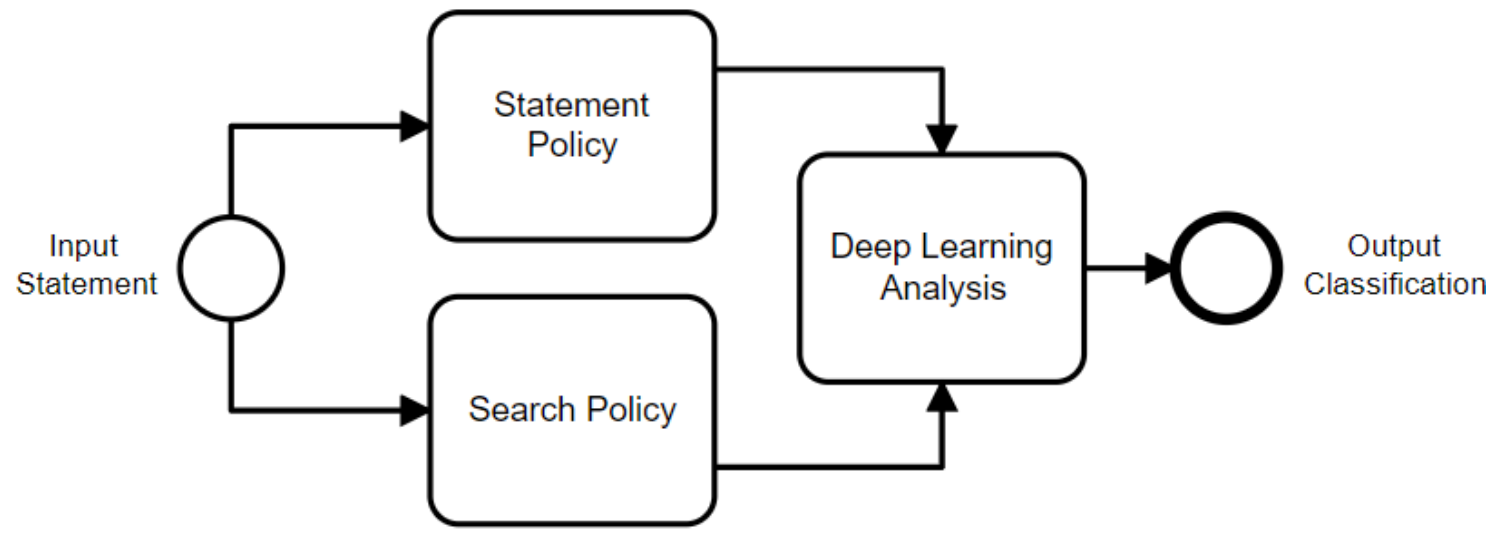

**Fig. 1.** Policies of the proposed framework.

The statement policy consists of a direct analysis of the initial statement, enabling MPSC to perform an 'offline' classification without considering any information obtained from external sources. In contrast, the search policy is an 'online' approach that provides additional information to be analyzed. The main subjects of the initial statement are retrieved and combined into a search query. This query is then provided to Application Programming Interfaces (APIs) of configurable news sources and search engines, obtaining recent news articles. A simultaneous search of multiple sources is performed by using the well-established News API, which aggregates their articles and ensures correctly formatted contents.

To create a search query, the framework relies on the Summa [18] and Yet Another Keyword Extractor (YAKE) [19] algorithms. Summa performs an unsupervised graph-based text summarization, creating a shorter and more concise statement, whereas YAKE performs an unsupervised keyword extraction, identifying and ranking the most relevant keywords within a statement. This approach enables the extraction of reliable queries for a more precise news search.

### 3.2 Analysis Architecture

In the deep learning analysis step, all obtained information is used to predict whether a statement is seemingly credible or suspicious. Lexical features are computed and provided to deep learning analysis layers. Then, their output is combined with additional syntactic features and analyzed in a final classification layer (see Fig. 2). This split architecture enables a more thorough analysis of a statement's legitimacy.

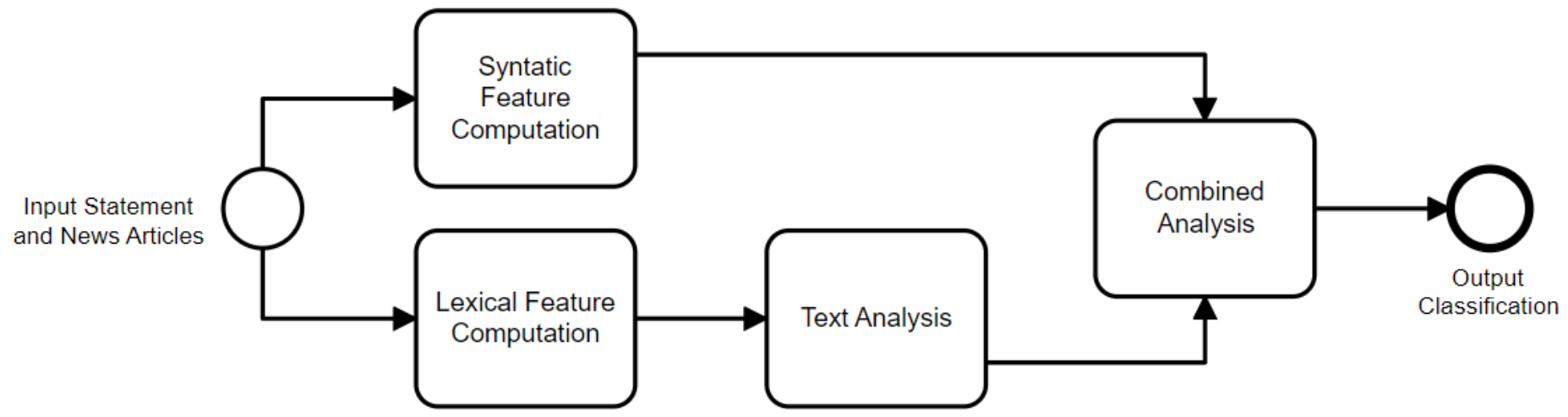

**Fig. 2.** Architecture of the proposed analysis.

For the computation of lexical features, all punctuation and stop words are removed from the text, and it is converted to lowercase. The text is then tokenized and lemmatized, being separated into smaller units that are then reduced to root words. Then, a word embedding is applied, representing individual words as real-valued vectors in a vector space. This leads to lexical features with a much smaller size than the entire utilized vocabulary, resulting in faster model training and predictions.

The databases used for tokenization and lemmatization were Treebank [20] and WordNet [21], respectively. The former contains sentence syntactic structure annotations, whereas the latter contains semantic relationships between words. The embedding was performed using WikiWords250, a token-based technique pre-trained on English Wikipedia corpus, based on a skip-gram version of Word2Vec with an out-of-vocabulary bucket [22].

Regarding the computation of syntactic features, it represents the main characteristics of a statement's structure as integer values. The features are calculated according to the following configuration: (i) total number of characters; (ii) number of uppercase letters; (iii) number of digits; (iv) number of punctuation marks; (v) number of unknown characters. These values are then normalized by removing the mean and scaling to unit variance, according to the data used to train a model.

## 4 Experimental Evaluation

A case study was conducted to evaluate the capabilities of the proposed framework and compare the performance of different deep learning models. The implementation relied on the Python programming language and the following libraries: *Numpy* and *Pandas* for data preprocessing and manipulation, *Nltk*, *Yake* and *Summa* for text processing, and *Tensorflow* for the deep learning models. The following subsections detail the most relevant aspects of the case study and present the obtained results.

### 4.1 Dataset and Data Preprocessing

To address the challenge of generalizing fake news detection to different subjects, without a classification model being overfit to specific types of statements, multiple datasets were considered. The ISOT [23] dataset contains real and fake news articles from legitimate and unreliable sources, which were marked as such by the well-established PolitiFact organization. Similarly, LIAR [24] and FakeNewsNet [25] are comprised of statements about distinct subjects directly fact-checked by PolitiFact. Even though their records are in different formats, they can be converted and merged into a single dataset. Considering that possibility, FNID [26] covers more recent subjects, compatible with both LIAR and FakeNewsNet formats.

The most adequate approach for ensuring a trustworthy case study was diversifying the utilized data by merging the four datasets. Therefore, a preprocessing stage was required before the merged dataset was usable. Besides discarding all other fields besides the text content and respective class label, the distinct labels of each dataset were converted to a single format, to be suitable for the binary classification of statements as credible or suspicious. Regarding the LIAR format, there were six classes that had to be aggregated. Due to the mostly false content of the other four classes, only "true" and "mostly-true" were considered credible.

After the label conversion, the holdout method was applied to randomly split the data into training, validation and evaluation sets, with 72669, 10501 and 10482 samples, respectively. The splitting was performed with stratification, preserving the original class proportions of each dataset. To provide an overview of the most common words of the final merged training set, a visual representation of word frequency was created for each class (see Fig. 3).

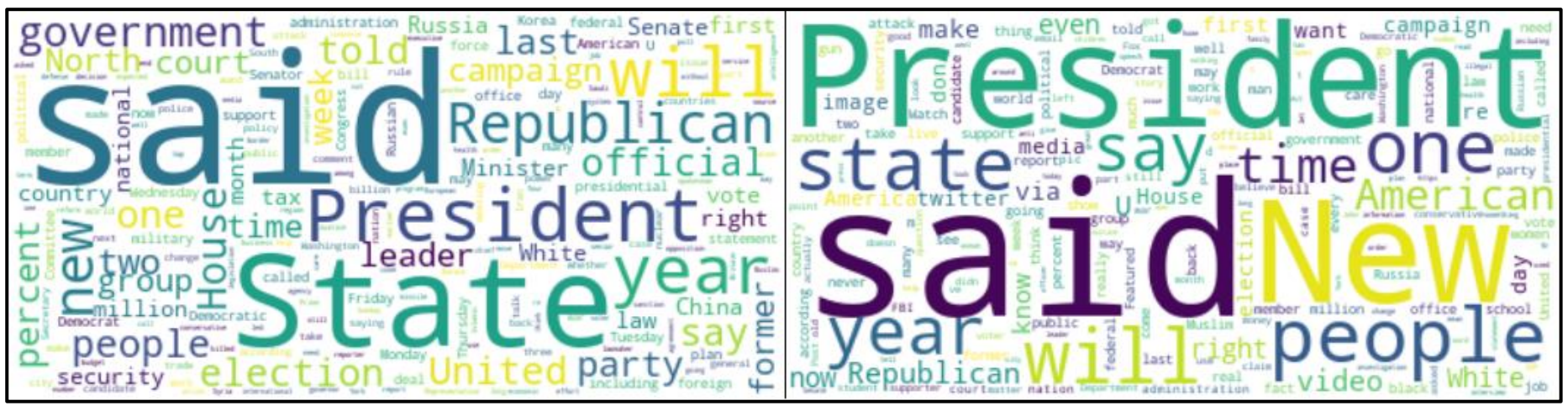

**Fig. 3.** Word frequency in credible (left) and suspicious (right) statements.

### 4.2 Evaluation Metrics

The performance of a classification model can be evaluated using the values reported by the confusion matrix. Considering credible statements as Negative and suspicious statements as Positive, it reports the number of True Positives (TP), True Negatives (TN), False Positives (FP) and False Negatives (FN). The considered metrics and their interpretation are described below [27][28].

Accuracy is a standard metric for classification tasks because it measures the proportion of correctly classified statements. However, a high accuracy score can be achieved even when a minority class is disregarded. For instance, if there is only 5% of fake news in an evaluation set, a model can fail to detect all of them, considering every sample as credible, and still achieve a score of 95%. Therefore, it is pertinent to analyze other metrics in order to provide a trustworthy evaluation.

Precision measures the proportion of predicted fake news that were actual fake, which indicates the relevance of a model's predictions. On the other hand, recall, which corresponds to the True Positive Rate, measures the proportion of actual fake news that were correctly predicted, reflecting a model's ability to detect suspicious statements. To provide a trustworthy score of a model's performance, a highly reliable metric is the F1-Score, also referred to as F-measure. It calculates the harmonic mean of precision and recall, considering both FP and FN. A high F1-Score indicates that suspicious statements are being correctly identified, while the number of credible statements falsely labeled as suspicious are kept at a minimum.

### 4.3 Models and Fine-tuning

Three distinct deep learning models were created: LSTM [29], GRU [30] and BERT [31]. The first two are RNNs that address the vanishing gradient problem by preserving past information that is considered relevant, which were trained from scratch. On the other hand, the latter is a state-of-the-art transformer that computes vector space representations of words. Since there are already well-established BERT models, a pre-trained model was further trained on the utilized dataset.

To account for the required architecture, the lexical features were provided to a model's layers. Then, their output was combined with the normalized syntactic features and analyzed in a final classification layer. Due to the high computational cost of training these deep learning models, they were fine-tuned through a Bayesian op-timization process [32]. It sought to minimize the cross-entropy loss obtained for the validation data, stopping the training when it stabilized.

The fine-tuning led to the use of the Adam algorithm with a learning rate of 0.001, avoiding a fast convergence to a suboptimal solution. A dropout rate of 0.2 was common to all models, inherently preventing overfitting by randomly ignoring 20% of the nodes during training. LSTM and GRU contained two layers with 256 and 128 nodes each, and a final classification layer with 32 nodes and the computationally efficient Rectified Linear Unit activation function. Due to the significantly higher number of nodes of the BERT layers, its final layer required 128 nodes, four times more than the other two models. The best validation losses were obtained with batch sizes of 32 for LSTM and 64 for both GRU and BERT.

### 4.4 Results and Discussion

The created deep learning models were applied to the proposed framework, using the evaluation set to perform a thorough performance comparison. Since LSTM, GRU and BERT can classify a statement using only its lexical features, the comparison included the results obtained when lexical and syntactic features were considered individually. Table 1 summarizes the case study results.

**Table 1.** Summary of obtained results.

| Model | Accuracy | Precision | Recall | F1-Score |
|---|---|---|---|---|
| Syntactic | 60.37 | 60.59 | 74.54 | 66.84 |
| LSTM | 90.41 | 91.39 | 90.74 | 91.06 |
| LSTM + Syntactic | 92.58 | **94.23** | 92.96 | **93.59** |
| GRU | 90.16 | 91.87 | 90.48 | 91.17 |
| GRU + Syntactic | 92.54 | 93.60 | 92.62 | 93.11 |
| BERT | 91.98 | 86.27 | 96.18 | 90.96 |
| BERT + Syntactic | **93.95** | 87.38 | **98.87** | 92.77 |

The results obtained by a model with only syntactic features evidence that these features are not reliable when used on a standalone basis, reaching an accuracy score of only 60.37%. On the other hand, the models using only lexical features, LSTM, GRU and BERT, achieved over 90% accuracy and a good balance between precision and recall. This noticeable difference confirms that is it necessary to analyze the actual text to achieve a good performance.

A trend arises when both lexical and syntactic features are used, with all three models obtaining approximately 2% higher scores than their individual approaches. The highest increase was obtained by GRU, from an accuracy score of 90.16% to 92.54%, but both LSTM and BERT behaved similarly. These stable increases demonstrate the effectiveness of combining both types of features for fake news detection.

Furthermore, the best accuracy and recall were obtained by the combined BERT, with both types of features. Despite the combined LSTM achieving the highest preci-sion and F1-Score, its recall was surpassed by the more complex BERT model. Therefore, the combined BERT identifies more false positives than LSTM, misclassifying a higher number of trustworthy news as suspicious, but is better at detecting fake news, in this case correctly identifying 98.87% of the suspicious statements.

## 5 Conclusions

This work addressed fake news detection by introducing MPSC, a framework that performs a multi-policy statement classification, considering both the statement itself and its related news articles. It analyses the legitimacy of each statement and predicts its level of suspiciousness, combining both lexical and syntactic features. A case study was conducted to evaluate the capabilities of the proposed framework, using four merged fake news detection datasets, and by training LSTM, GRU and BERT models to perform the classification.

The obtained results demonstrate that MPSC can be significantly advantageous to the fake news identification process, when compared to the costs inherent to the traditional approach of professional analysis. The introduced framework can be used to automatically analyze large amounts of news, as it quickly and accurately identifies suspicious statements, or it could help professional fact-checkers to mark dubious news for further analysis. The best accuracy and recall were obtained by BERT with syntactic features, although LSTM with syntactic features achieved a good balance between precision and recall, exhibited by its F1-Score.

In the future, additional policies can be added to this framework, to provide a more thorough analysis of a statement's context. For instance, the similarity of a statement to sentences present in news articles from a configured list of trusted sources could also be utilized to assess its credibility. It is also pertinent to more comprehensive datasets, covering broader subjects and containing greater quantities of labeled statements. Future analyses could also include the degree of similarity between a statement and its related news articles, further improving fake news detection.

**Acknowledgments.** The present work was partially supported by the Norte Portugal Regional Operational Programme (NORTE 2020), under the PORTUGAL 2020 Partnership Agreement, through the European Regional Development Fund (ERDF), within project "Cybers SeC IP" (NORTE-01-0145-FEDER-000044). This work has also received funding from UIDB/00760/2020.